%% file: 08072-source.tex
\documentclass[runningheads]{llncs}

\usepackage{eccv}

\usepackage{eccvabbrv}
\usepackage{marvosym}
\usepackage{graphicx}
\usepackage{booktabs}

\usepackage[accsupp]{axessibility}  
\usepackage{algorithm}
\usepackage{algpseudocode}
\usepackage{wrapfig}

\usepackage{multirow}
\usepackage[normalem]{ulem}
\useunder{\uline}{\ul}{}

\usepackage{hyperref}

\usepackage{orcidlink}
\makeatletter
\def\thanks#1{\protected@xdef\@thanks{\@thanks
        \protect\footnotetext{#1}}}
\makeatother

\begin{document}

\title{Diagnosing and Re-learning for Balanced Multimodal Learning} 


\author{Yake Wei\inst{1} \and
Siwei Li\inst{2}\and
Ruoxuan Feng\inst{1} \and
Di Hu\textsuperscript{\Letter}\inst{1,3}
\thanks{\textsuperscript{\Letter}Corresponding author.}}

\authorrunning{Y.~Wei et al.}

\institute{Gaoling School of Artificial Intelligence, Renmin University of China, China \\
\email{\{yakewei,fengruoxuan,dihu\}@ruc.edu.cn} \and 
Department of Electronic Engineering, Tsinghua University, China  \\
\email{lisw19@mails.tsinghua.edu.cn}  \and
Engineering Research Center of Next-Generation Search and Recommendation}

\maketitle

\input{content/abstract}

\input{content/introduction}

\input{content/relatedwork}

\input{content/method}

\input{content/experiment}

\input{content/conclusion}


\section*{Acknowledgements}
This research was supported by National Natural Science Foundation of China (NO.62106272), and Public Computing Cloud, Renmin University of China.

%
%
\bibliographystyle{splncs04}
\bibliography{main}
\end{document}

%% file: content/abstract.tex
\begin{abstract}
To overcome the imbalanced multimodal learning problem, where models prefer the training of specific modalities, existing methods propose to control the training of uni-modal encoders from different perspectives, taking the inter-modal performance discrepancy as the basis. However, the intrinsic limitation of modality capacity is ignored. The scarcely informative modalities can be recognized as ``worse-learnt'' ones, which could force the model to memorize more noise, counterproductively affecting the multimodal model ability. Moreover, the current modality modulation methods narrowly concentrate on selected worse-learnt modalities, even suppressing the training of others. Hence, it is essential to consider the intrinsic limitation of modality capacity and take all modalities into account during balancing. To this end, we propose the Diagnosing \& Re-learning method. The learning state of each modality is firstly estimated based on the separability of its uni-modal representation space, and then used to softly re-initialize the corresponding uni-modal encoder. In this way, the over-emphasizing of scarcely informative modalities is avoided. In addition, encoders of worse-learnt modalities are enhanced, simultaneously avoiding the over-training of other modalities. Accordingly, multimodal learning is effectively balanced and enhanced. Experiments covering multiple types of modalities and multimodal frameworks demonstrate the superior performance of our simple-yet-effective method for balanced multimodal learning. The source code and dataset are available at \url{https://github.com/GeWu-Lab/Diagnosing_Relearning_ECCV2024}.

\keywords{Multimodal learning \and Learning state diagnosing \and Re-learning}

\end{abstract}

%% file: content/introduction.tex
\section{Introduction}

Inspired by the human's multi-sensory perception, multimodal learning where information from diverse sensors is jointly utilized, has witnessed tremendous progress in recent years~\cite {baltruvsaitis2018multimodal,liang2022foundations}. 
Even with current developments, the question of how to assess and facilitate learning of individual modality remains open in the multimodal learning field. Especially, recent studies have found that some modalities in the multimodal model are less learnt than others~\cite{huang2022modality,peng2022balanced}, called the imbalanced multimodal learning problem. This imbalance in modality utilization hinders the potential of multimodal learning, and even could make the multimodal model fail its uni-modal counterpart~\cite{wang2020makes,peng2022balanced}. Accordingly, a series of empirical methods are proposed to \emph{balance} the uni-modal learning and achieve better multimodal performance~\cite{peng2022balanced,wu2022characterizing,fan2023pmr,li2023boosting}. During the uni-modal balancing process, two keys are selection basis and balancing strategy. 

In existing methods~\cite{peng2022balanced,wu2022characterizing,fan2023pmr,li2023boosting}, it is commonly believed that the modality with better prediction performance is the ``well-learnt'' modality, and correspondingly, the other ``worse-learnt'' modalities are the ones that need to be trained emphatically during uni-modal balancing. However, \textbf{they ignore the intrinsic limitation of modality capacity, where some modalities naturally have scarcely label-related information and more noise}. For cases of these modalities, the limited information causes their limited prediction performance, not just insufficient training. Although with worse prediction performance, purely emphasizing the training of these modalities could not bring many additional benefits and even force the model to memorize more noise, affecting the model ability. To further illustrate this problem, we modify the audio-vision CREMA-D dataset, and add white Gaussian noise into its audio data, making the audio modality with limited discriminative information but numerous noise. As~\autoref{fig-teaser_noise}, all existing imbalanced multimodal learning methods experience a performance drop compared with joint-training baseline. This phenomenon verifies that they wrongly push the training of scarcely informative modality with intrinsic limitation, counterproductively making them lose efficacy.

\input{figure_table/figure_teaser}

Upon the design of the balancing strategy, \textbf{existing methods narrowly concentrate on the learning of selected worse-learnt modalities~\cite{peng2022balanced,wu2022characterizing,fan2023pmr,li2023boosting}.} Some even disturb the training of well-learnt modality~\cite{peng2022balanced,li2023boosting}, to facilitate the training of others. Inevitably, the ignorance or even suppression of well-learnt modality potentially affects its learning. As~\autoref{fig-teaser_cd} and~\autoref{fig-teaser_ks}, in existing imbalanced methods, although improving multimodal performance, the quality of the well-learnt audio modality can be worse than the joint-training baseline, especially on the Kinetics Sounds dataset. Recognizing these limitations, the challenge lies in how to overcome scarcely informative modalities cases and take all modalities into account during balancing. 

In this paper, we propose the Diagnosing \& Re-learning method, which periodically softly re-initialize the uni-modal encoder with representation separability as the basis. Since many multimodal models only have one multimodal output, it is hard to directly obtain the uni-modal learning state without additional modules. To this end, we focus on the uni-modal representation space, which is easier to access and can indirectly reflect the modality discriminative ability~\cite{sehwag2020separability}. Concretely, the separability of train-representation and validation-representation are assessed by clustering, and utilized to diagnose the uni-modal learning state. In this way, the learning of each modality is well estimated individually. Then, to balance the uni-modal training, uni-modal encoders are softly re-initialized based on their learning state. \textbf{For well-learnt modalities}, their encoder has a greater re-initialization strength. This strategy helps the model reduce the reliance on them, and enhance the learning of other still under-fitting modalities. Simultaneously, the greater re-initialization of well-learnt modality avoids its over-training, even potentially improving the generalization~\cite{ash2020warm,zaidi2023does}. \textbf{For worse-learnt modalities}, their encoders are slightly re-initialized, which is also beneficial for escaping memorizing data noise that harms generalization. 
\textbf{When one modality is scarcely informative}, our method will not wrongly over-emphasize its training, and the re-initialization for its encoder helps encoders avoiding memorize data noise. Therefore, \textbf{our strategy can benefit all modalities at the same time}. In addition, the soft re-initialization partially preserves previously learnt knowledge already accrued by the network~\cite{ash2020warm,sokar2023dormant}. It safeguards the collaboration between modalities, ensuring that the collaborative knowledge is not completely discarded but rather fine-tuned. 

Based on~\autoref{fig-teaser_noise}, our method can well handle the scarcely informative modality case and ideally achieves performance improvement. Moreover, as~\autoref{fig-teaser_cd} and~\autoref{fig-teaser_ks}, it also effectively enhances the learning of all modalities. Our method is flexible and can be equipped with diverse multimodal frameworks, including the multimodal Transformer. Overall, our contributions are three-fold. \textbf{Firstly,} we point out that existing imbalanced multimodal learning methods often ignore the intrinsic limitation of modality capacity and the well-learnt modality during balancing. \textbf{Secondly,} we propose the Diagnosing \& Re-learning method to well balance uni-modal training by softly re-initialize encoders based on the uni-modal learning state. \textbf{Thirdly,} experiments across different types of modalities and multimodal frameworks substantiate the superior performance of our simple-yet-effective method.

%% file: figure_table/figure_teaser.tex
\begin{figure}[t]
\centering
	\begin{subfigure}[t]{0.3\linewidth}
			\centering
			\includegraphics[width=\textwidth]{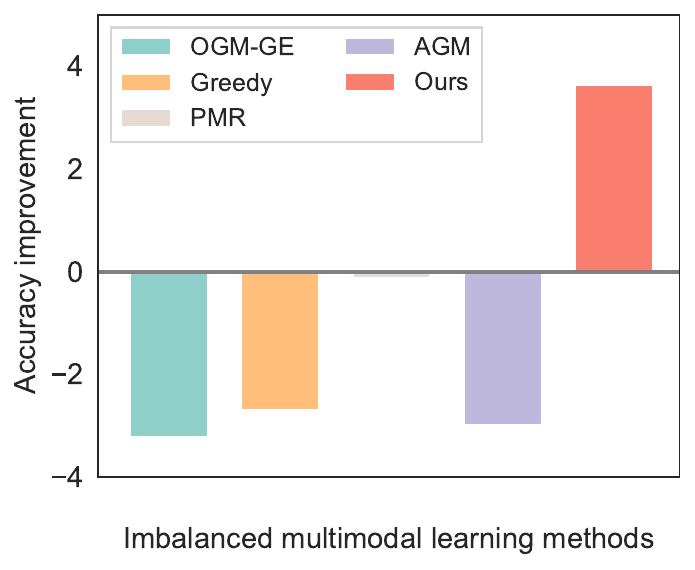}
			\caption{Scarcely informative case.}
   \label{fig-teaser_noise}    
	\end{subfigure}
    \begin{subfigure}[t]{0.3\linewidth}
			\centering
			\includegraphics[width=\textwidth]{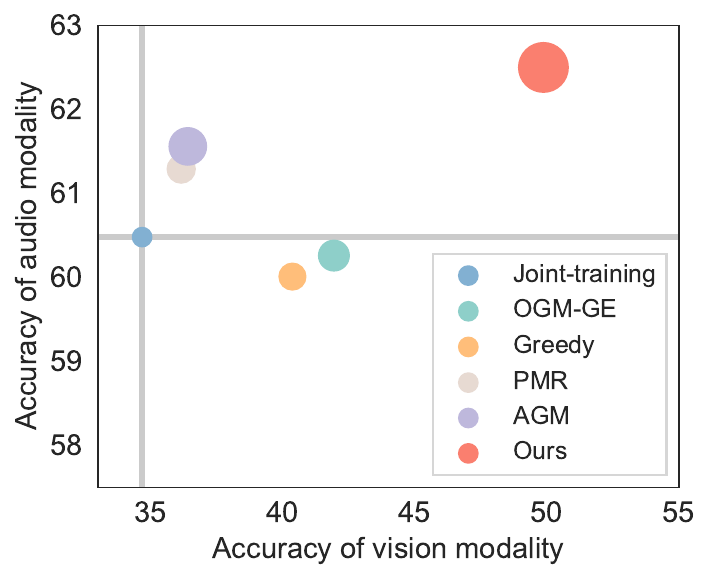}
			\caption{CREMA-D dataset.}
			\label{fig-teaser_cd}
	\end{subfigure}
	\begin{subfigure}[t]{0.3\linewidth}
			\centering
			\includegraphics[width=\textwidth]{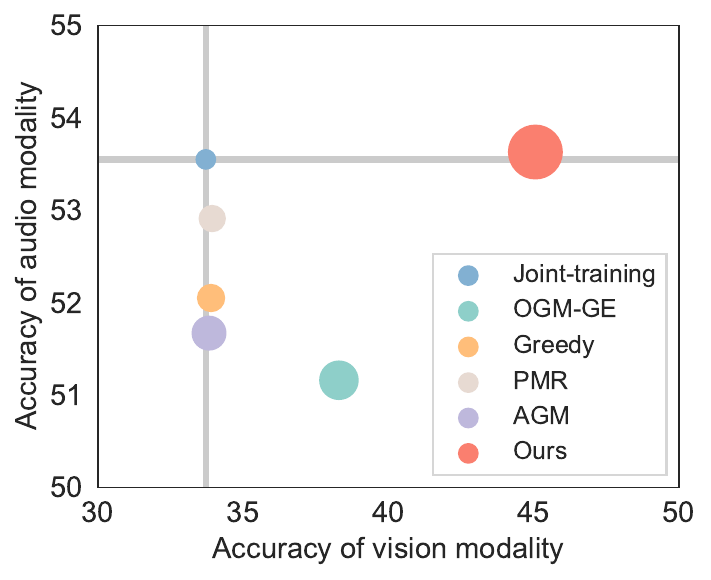}
			\caption{Kinetics Sounds dataset.}
   \label{fig-teaser_ks}
	\end{subfigure}
    \caption{\textbf{(a): Scarcely informative modality case.} It shows the accuracy improvement compared with the joint-training baseline. Only our method has a positive performance improvement. \textbf{(b)\&(c): Uni-modal encoder quality evaluation and comparison.} The uni-modal evaluation (Acc audio and Acc vision) is obtained by fine-tuning a new uni-modal classifier with the corresponding trained uni-modal encoder. A larger spot size reflects a better multimodal performance. Our method is superior in both multimodal performance and all uni-modal performance.}
    \label{fig:teaser}
\end{figure}

%% file: content/relatedwork.tex
\section{Related Work}

\textbf{Multimodal learning.}
Motivated by the multi-sensory experiences of humans, the field of multimodal learning has gained significant attention and experienced rapid growth in recent years~\cite{liang2022foundations}. Multimodal learning involves the development of models capable of simultaneously integrating information from various modalities. Research in multimodal learning spans diverse domains, including such as multimodal recognition~\cite{xu2023multimodal,yadav2021review} and audio-visual scene understanding~\cite{zhu2021deep,wei2022learning}. Commonly employed multimodal frameworks typically entail the extraction and fusion of uni-modal features, followed by the optimization of all modalities with joint learning objectives. However, besides the superficial multimodal performance across various tasks, the inherent learning of different modalities remains under-explored. \\

\noindent \textbf{Imbalanced multimodal learning.}
Recent studies have revealed the imbalanced multimodal learning problem, where models prefer certain modalities over others, limiting their overall effectiveness~\cite{huang2022modality,peng2022balanced}. A variety of strategies have been suggested, focusing on balancing the optimization of individual modalities~\cite{wu2022characterizing,fan2023pmr,li2023boosting,wei2024innocent,wei2024enhancing,yang2024Quantifying}. For example, Peng~\emph{et al.}~\cite{peng2022balanced} proposed gradient modulation strategy, which dynamically monitors the contribution difference of various modalities to the final prediction during training and mitigates the gradient magnitude of dominant modality to focus more on other modalities. However, in these studies, they ignore the intrinsic limitation of modality capacity. One modality can be naturally scarcely informative and with plenty of noise. In addition, in these methods, when balancing uni-modal learning, only the selected worse-learnt modality is focused on. Some of them even intentionally impair the training of well-learned modalities to promote the learning of others~\cite{peng2022balanced,li2023boosting}. In this paper, we first diagnose the learning state of individual modality by its own representation separability without additional modules. Then, the learning state is used as the basis of soft encoder re-initialization to encourage further learning of under-fitting modalities while concurrently preventing the over-training of originally well-learnt modalities and scarcely informative modalities. It ensures better learning of all modalities. \\

\noindent \textbf{Network re-initialization.}
Recent studies suggest that re-initializing the network parameters during training can effectively improve model performance and parameter utilization~\cite{ash2020warm,alabdulmohsin2021impact,zaidi2023does,sokar2023dormant}. These methods involve re-initializing and transforming a part or all of the parameters of a neural network periodically. For instance, Alabdulmohsin~\emph{et al.}~\cite{alabdulmohsin2021impact} proposed the Layer-wise Re-initialization strategy, which re-initializes the architecture block-by-block during training. Qiao~\emph{et al.}~\cite{qiao2019neural} proposed to detect unsatisfactory components in a neural network and re-initialize them to encourage them can better fit the tasks. Here we introduce the idea of network re-initialization, but with a very different intention. The uni-modal encoder is re-initialized based on its learning state, to ensure the benefits of all modalities as well as balancing uni-modal training.

%% file: content/method.tex
\section{Method}
\subsection{Framework and notations.}
\textbf{Multimodal framework.} As the left part of~\autoref{fig:method}, data of each modality is firstly fed into the corresponding uni-modal encoder to extract features. Then these uni-modal features are fused to obtain the multimodal feature. Our method has no reliance on the multimodal fusion strategy, and can cover simple fusion methods (\emph{e.g.,} concatenation), and complex fusion methods (\emph{e.g.,} cross-modal interaction). The fused feature is fed into the final multimodal classifier. One multimodal loss, cross-entropy, is utilized to optimize the model.

\noindent \textbf{Notations.} For the dataset with $K$ modalities, the training set is denoted as $\mathcal{D}$ with $N_{\mathcal{D}}$ samples and the validation set is denoted as $\mathcal{V}$ with $N_{\mathcal{V}}$ samples. Each data sample $x=\{x^1,x^2,\cdots,x^K\}$ is with $K$ modalities. The category number of the dataset is $M$. For each modality $k$, where $k \in \{1,2,\cdots, K\}$, parameters of its encoder are denoted as $\theta_k$. $\theta_k^{\text{init}}$ represents the initialized parameter value.

\input{figure_table/figure_method}
\subsection{Diagnosing: uni-modal learning state estimation}
\label{sec:clustering}
In multimodal learning, many multimodal models only have one multimodal output. Therefore, it is hard to directly obtain the uni-modal learning state without additional modules. In former studies, the estimation of uni-modal learning state often relies on specific fusion strategy~\cite{peng2022balanced,fan2023pmr}. This limits their application to a wider range of scenarios. Elaborately designing ways to obtain uni-modal output is clearly complicated and not universal, since the multimodal fusion strategies are diverse. To well diagnose the uni-modal learning state without any additional modules or reliance on fusion strategies, we propose to focus on the uni-modal representation space. It is known that the separability can reflect the representation quality~\cite{sehwag2020separability}. Observing and comparing the separability of each extracted uni-modal representation is promising to capture the learning state. To evaluate representation separability, one straightforward idea is k-means clustering~\cite{macqueen1967some}. 

For data sample $x_i$ in the training set $\mathcal{D}$ with $N_{\mathcal{D}}$ samples, its $k-$th uni-modal feature that extracted by its $k-$th encoder $\theta_k$ is: $\phi^k_i=\theta_k(x^k_i)$. Then, to evaluate the separability of uni-modal features, it needs to split the set of all $k-$th uni-modal training features, $\Phi^k_{\mathcal{D}}=  \{ \phi^k_1, \phi^k_2,\cdots, \phi^k_{N_{\mathcal{D}}} \}$, into $M$ clusters. The set of all clusters is $\textbf{C}=\{ C_1, C_2, \cdots, C_M\}$, where $M$ is the category number. 

Concretely, when splitting uni-modal features into clusters, $M$ samples in $\Phi^k_{\mathcal{D}}$ is firstly randomly picked as the centroid of $M$ clusters. Then, at the \textbf{assignment step}, each sample is assigned to the cluster with the nearest mean based on Euclidean distance. Concretely, sample $\phi^k_i$ is assigned to $m-$th cluster $C_m$ with centroid $O_m$  when:
\begin{equation}
     \left \| \phi^k_i - O_m \right \|^2 \le \left \| \phi^k_i- O_j \right \|^2 \ \forall j, 1 \le j \le M.
\end{equation}
$\| \cdot \|$ denotes the $L_2$-norm. After that, at the \textbf{updating step}, the centroid of each cluster is recalculated based on the current cluster:
\begin{equation}
   O_m = \frac{1}{\left|C_m\right|} \sum_{\phi^k_i \in C_m} \phi^k_i. 
\end{equation}

After a given number of iterations between the assignment step and the updating step or the assignments no longer change, we have the final clustering results. For high-quality uni-modal representation, its ideal separability of feature space will bring satisfied clustering results. To evaluate the clustering results, we consider the clustering purity, which is a representative measurement for clustering quality~\cite{wong2015short}. Concretely, we first divide samples in $\Phi^k_{\mathcal{D}}$ into $M$ groups based on the ground truth labels and have classification sets $\textbf{Z}=\{ Z_1, Z_2, \cdots, Z_M \}$. Comparing the former clustering sets $\textbf{C}$ and classification sets $\textbf{Z}$, the purity is
\begin{equation}
\label{equ:purity}
 P^k_{\mathcal{D}}=   \frac{1}{N_{\mathcal{D}}}\sum_{C_m\in \textbf{C}}\max_{Z_m\in \textbf{Z}}{|C_m \cap Z_m|}.
\end{equation}
It reflects the extent to which clusters contain a single class. Higher purity means better clustering results. And the uni-modal representation is of higher quality. 

To diagnose the learning state of modality $k$, comparing the representation quality discrepancy between the training set $\mathcal{D}$ and the validation set $\mathcal{V}$ would be a useful reference. We know that when one model is well-learnt or even over-trained, its validation performance would be not increased according to the training performance~\cite{ying2019overview}. This can also happen in the uni-modal encoder, bringing a gap between their train and validation representation quality. This gap is expected to reflect the learning state of one modality. Concretely, for the validation set $\mathcal{V}$, we also conduct the clustering algorithm and obtain its purity $P^k_{\mathcal{V}}$. And the gap between training set purity $P^k_{\mathcal{D}}$ and validation set purity $P^k_{\mathcal{V}}$ is:
\begin{equation}
\label{equ:gap}
    g^k = | P^k_{\mathcal{D}} - P^k_{\mathcal{V}} |.
\end{equation}
Based on the property of purity, $g^k \in [0,1]$. This purity gap reflects the quality gap between train and validation representation (Observations about this gap are provided in~\autoref{sec:purity_exp}). When the value of purity gap $g^k$ is large, this modality is well-learnt or even over-trained. In this way, the learning state of one modality is diagnosed individually.

\input{figure_table/alg}

\subsection{Re-learning: uni-modal re-initialization based on learning state}
In~\autoref{sec:clustering}, the uni-modal learning state is diagnosed by the separability discrepancy between training and validation representation space. Then, to balance the uni-modal training, we propose to softly re-initialize all uni-modal encoders based on their diagnosed learning state. This re-initialization breaks the model's reliance on one specific modality, and potentially enhances the model's generalization ability by re-learning multimodal data. Specifically, the re-initialization strength $\alpha_k$ for modality $k$ is calculated based on purity gap:
\begin{equation}
    \alpha_k = \text{tanh}(\lambda \cdot g^k),
    \label{equ:strength}
\end{equation}
where $\lambda >1$ is the hyper-parameter to further control the re-initialization strength. Then we can have $\lambda \cdot g^k \geq 0$ and $\alpha_k \in [0,1)$. The use of function $\text{tanh(}x\text{)}$\footnote{Other functions that satisfy these properties can also be used. We provide more experiments in the supplementary material.} aims to map the final re-initialization strength to a value between $0$ and $1$, while ensuring a monotonically increasing property when $x\geq 0$. These properties make the re-initialization strength $\alpha_k$ proportional to the purity gap $g^k$. Then, the encoder parameters of modality $k$ are re-initialized by:
\begin{equation}
\label{equ:reinitialize}
    \theta_k= (1-\alpha_k) \cdot \theta_k^{\text{current}} + \alpha_k \cdot \theta_k^{\text{init}},
\end{equation}
where $\theta_k^{\text{current}}$ is the current parameter and $\theta_k^{\text{init}}$ is the initialized parameter. 

With our strategy, on the one hand, for the well-learnt modalities, its encoder experiences a greater re-initialization, which effectively makes the model temporarily get rid of the dependence on them and enhances the learning of other still under-fitting modalities. Meanwhile, after re-initialization, the model would re-learn the former well-learnt data. This process can help to prevent the model from confidently fitting to the noise, avoiding over-training for well-learnt modalities. On the other hand, for other modalities (even they are scarcely informative), the slight re-initialization in their encoder also helps to prevent the memorization of data noise that negatively affects generalization. Overall, our method can benefit all modalities simultaneously. In addition, our soft re-initialization maintains a portion of the knowledge previously acquired by the model. It protects the learnt inter-modal correlation to some extent, making sure that the shared knowledge is not entirely lost but instead refined. The proposed method is illustrated in~\autoref{fig:method}, and the entire training process is shown in~\autoref{alg:alg}. The Diagnosing \& Re-learning strategy is conducted every $H$ epoch.

%% file: figure_table/figure_method.tex
\begin{figure*}[t]
    \centering
    \includegraphics[width=1\linewidth]{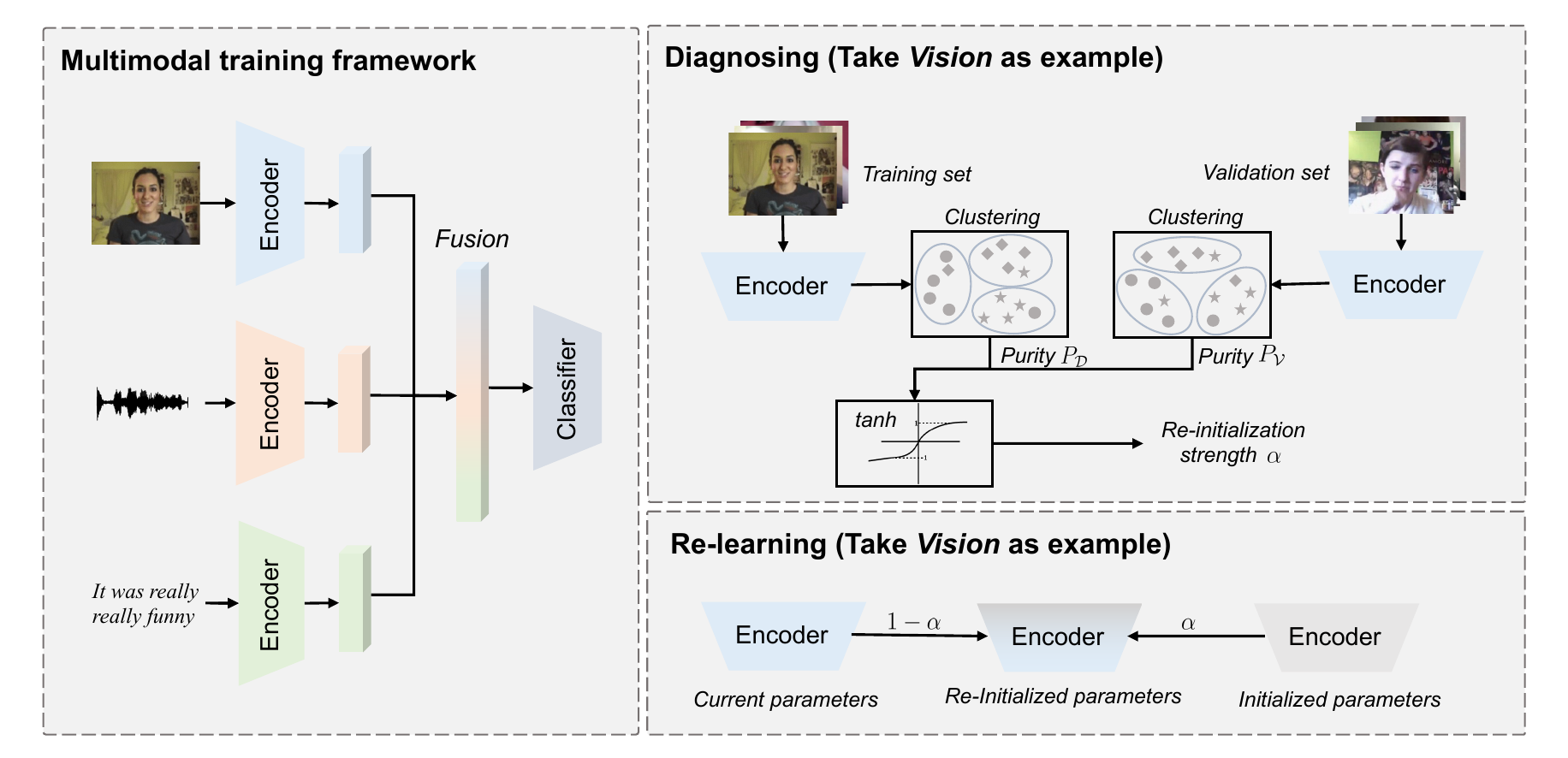}
    \caption{Illustration of multimodal framework and the proposed Diagnosing \& Re-learning method.}
    \label{fig:method}
\end{figure*}

%% file: figure_table/alg.tex
\begin{algorithm}[t]
\caption{Diagnosing \& Re-learning}
\label{alg:alg}
\begin{algorithmic}
\Require Training set $\mathcal{D}$, validation set $\mathcal{V}$, epoch number $T$, uni-modal encoder parameters $\theta_{k}$, $k \in \{1,2,\cdots,K\}$, Diagnosing \& Re-learning frequency $H$.
\For{$t=0,\cdots,T-1$} 
    \State Train and update parameters;
    \If{$t \mod H == 0$}
        \For{$k=1,\cdots,K$}
            \State Extract training feature set $\Phi^k_{\mathcal{D}}$ and validation feature set $\Phi^k_{\mathcal{V}}$;
            \State Conduct clustering algorithm on $\Phi^k_{\mathcal{D}}$ and have its purity $P^k_{\mathcal{D}}$;
            \State Conduct clustering algorithm on $\Phi^k_{\mathcal{V}}$ and have its purity $P^k_{\mathcal{V}}$;
            \State Calculate the purity gap $g^k$ based on~\autoref{equ:gap};
            \State Calculate re-initialization strength $\alpha_k $ based on~\autoref{equ:strength};
            \State Reinitialize encoder parameters $\theta_{k}$ with $\alpha_k $ based on~\autoref{equ:reinitialize}. 
        \EndFor
    \EndIf
\EndFor
\end{algorithmic}
\end{algorithm}

%% file: content/experiment.tex
\section{Experiment}

\subsection{Dataset}

\noindent \textbf{CREMA-D}~\cite{cao2014crema} is an emotion recognition dataset with two modalities, audio and vision. This dataset covers six emotions: angry, happy, sad, neutral, discarding, disgust and fear. The whole dataset contains 7442 clips.

\noindent \textbf{Kinetic Sounds}~\cite{arandjelovic2017look} is an action recognition dataset with two modalities, audio and vision. This dataset contains 31 human action classes, which are selected from the Kinetics dataset~\cite{kay2017kinetics}. It contains 19k 10-second video clips.

\noindent \textbf{UCF-101}~\cite{soomro2012ucf101} is an action recognition dataset with two modalities, RGB and optical flow. This dataset contains 101 categories of human actions. The entire dataset is divided into a 9,537-sample training set and a 3,783-sample test set according to the original setting.

\noindent \textbf{CMU-MOSI}~\cite{zadeh2016mosi} is a sentiment analysis dataset with three modalities, audio, vision and text. It is annotated with utterance-level sentiment labels. This dataset consists of 93 movie review videos segmented into 2,199 utterances.

\subsection{Experimental settings}
For the CREMA-D and the Kinetic Sounds dataset, ResNet-18 is used as the backbone and models are trained from scratch. For the UCF-101 dataset, ResNet-18 is also used as the backbone and is ImageNet pre-trained. For the CMU-MOSI dataset, transformer-based networks are used as the backbone~\cite{liang2021multibench} and the model is trained from scratch. The choices of architecture and initialization follow former imbalanced multimodal learning studies, to have a fair comparison. During training, we use the SGD optimizer with momentum ($0.9$) and set the learning rate at $1e-3$. All models are trained on 2 NVIDIA RTX 3090 (Ti). 

In experiments, our method is conducted every $5$ epoch for the CMU-MOSI dataset and every $20$ epoch for others. $\lambda$ is $3$, $7$, $7$, $8$ for CREMA-D, Kinetics Sounds, UCF-101 and CMU-MOSI dataset respectively. More ablation studies and comparisons are provided in the supplementary material.

\input{figure_table/table_imbalance}

\input{figure_table/table_gb}

\subsection{Comparison with imbalanced multimodal learning methods}
To assess the efficacy of our method in addressing the imbalanced multimodal learning problem, we conduct comparisons with recent studies: \textbf{G-Blending~\cite{wang2020makes}, OGM-GE~\cite{peng2022balanced}, Greedy~\cite{wu2022characterizing}, PMR~\cite{fan2023pmr} and AGM~\cite{li2023boosting}. } Joint-training is the widely used baseline for the imbalanced multimodal learning problem, with concatenation fusion and one multimodal cross-entropy loss function~\cite{peng2022balanced,fan2023pmr,li2023boosting}. The results are shown in~\autoref{tab:imbalance}. We provide experiments across several datasets with different modalities, like audio, vision and optical flow. Based on the results, we first observe that all these imbalanced multimodal learning methods achieve improvement in the multimodal performance, exhibiting the existence of the imbalanced multimodal learning problem and the necessity of balancing uni-modal learning during training. More than that, our method consistently exhibits superior performance across multiple datasets with different types of modalities, outperforming other methods. This demonstrates the effectiveness of our Diagnosing and Re-learning strategy, which takes all modalities into account.

In~\autoref{tab:imbalance}, the G-Blending method also achieves considerable improvement in the model performance, especially on the Kinetics Sounds and UCF-101 datasets. However, compared with other methods that only have one multimodal joint loss, it introduces additional uni-modal classifier and correspondingly uni-modal loss functions. These additional modules are definitely helpful for controlling the training of individual modalities. As stated before, our method is flexible and not limited to special multimodal frameworks. Therefore, to have a fair comparison, we also introduce the same uni-modal classifier and loss function as G-Blending. The results are shown in~\autoref{tab:gb}. After introducing the same uni-modal classifier and loss function, our method (Ours\dag \ in the table) significantly outperforms the G-Blending method. In addition, these results also suggest that our method of targeted uni-modal encoder re-initialization on the basis of learning state can be effectively integrated with other modules while maintaining its effectiveness.

\input{figure_table/table_three}

\subsection{Comparison in more general multimodal frameworks}
In multimodal learning, besides the widely used late-fusion framework with the convolutional neural network backbone, more complex transformer-based backbone and cross-modal interaction modules also have a wide range of applications. In this section, we conduct a comparison of the transformer-based multimodal frameworks. Results are shown in~\autoref{tab:three}. For the CMU-MOSI dataset, Transformer is used as the backbone, following~\cite{liang2021multibench} and the model is trained from scratch. For the CREMA-D dataset, experiments in this section use the representative multimodal Transformer backbone, MBT~\cite{nagrani2021attention}. It has both single-modal layers and cross-modal interaction layers with dense cross-modal interaction modules. In experiments, the model is ImageNet pre-trained. The results lead us to make the following observation that existing imbalanced multimodal learning methods may lose efficacy in Transformer-based frameworks involving cross-modal interactions. For instance, the Greedy method~\cite{wu2022characterizing} is even inferior to the joint-training baseline on the CREMA-D dataset. Conversely, our method, which has no reliance on special types of multimodal frameworks, demonstrates its ideal versatility and superior performance.

\subsection{Comparison in more-than-two modality case}
In current imbalanced multimodal learning methods, many of them only focus on the case of two modalities~\cite{wu2022characterizing,peng2022balanced,fan2023pmr}. This limitation greatly hampers their application in broader scenarios. In contrast, our method has no restriction on the number of modalities. In this section, we compare these methods in the more-than-two modality case, on the CMU-MOSI dataset with three modalities: audio, vision and text. Among existing imbalanced multimodal methods, the Greedy~\cite{wu2022characterizing} method could not be extended to this case suitably. In the original paper of OGM-GE~\cite{peng2022balanced} and PMR~\cite{fan2023pmr}, they only provide methods for two modalities. Hence, to have a sufficient comparison, we retain the core uni-modal balancing strategy of OGM-GE and PMR, and extend them to more than two modality cases. Based on results in the left part of~\autoref{tab:three}, our method is not limited by the number of modalities and remains effective in this case.

\input{figure_table/table_noise}

\subsection{Comparison in scarcely informative modality case}
As analyzed before, in existing imbalanced multimodal methods, the prevailing view is that the modality exhibiting superior predictive ability is considered the ``well-learnt'' modality, while the remaining ``worse-learnt'' modalities require additional training during uni-modal balancing. Therefore, the existing imbalanced multimodal methods will fail when facing the case that modalities naturally have scarcely label-related information and more noise. This kind of scarcely informative modality will be recognized as the ``worse-learnt'' during training, and these methods will explicitly enhance their learning. However, enhancing their training offers no extra advancement and may even prompt the model to memorize more noise, affecting its effectiveness. 

To validate this problem, we consider the scarcely informative modality case. We modify the audio data of the CREMA-D dataset, adding extra white Gaussian noise to make it noisier and scarcely discriminative. Based on the results shown in~\autoref{tab:noise}, all these imbalanced multimodal methods suffer a decline in performance when compared to the joint-training baseline, even the G-Blending~\cite{wang2020makes} method with additional uni-modal modules and learning objectives. But our method continues to secure significant enhancement in this challenging scarcely informative modality case. The reason could be that this scarcely informative modality is often with both low-quality training representation and validation representation, due to the existence of much irrelevant noise. It has a small purity gap. Then, its encoder will be re-initialized with a slight percentage with our method, which is helpful to avoid the over-learning of the noisy data, even potentially improving model generalization~\cite{ash2020warm,alabdulmohsin2021impact,zaidi2023does}. 
Hence our method can well handle the scarcely informative modality case. In addition, there are related cases where one modality is not scarcely informative, but it is still noticeably less informative than others. For example, in the UCF-101 dataset, the accuracy of the individually-trained optical flow model is $58.9$, and the individually-trained RGB model is $73.2$. Our method also maintains superior, as shown in~\autoref{tab:imbalance}.

\input{figure_table/figure_tsne}

\subsection{Uni-modal representation quality analysis}
Beyond the comparison in overall multimodal performance, we also evaluate the uni-modal representation quality of our method to comprehensively reflect how well the imbalanced multimodal learning method is addressed. In terms of quantitative analysis, we fine-tune a new uni-modal classifier for the trained uni-modal encoder. Results are shown in~\autoref{fig-teaser_cd} and~\autoref{fig-teaser_ks}. Different from other imbalanced multimodal learning methods that ignore or disrupt the training of well-learnt modalities, our method demonstrates an ideal enhancement for all modalities. In addition to the quantitative analysis, we also performed a qualitative analysis of uni-modal representation. As shown in~\autoref{fig:tsne}, we visualize the uni-modal representation by t-SNE~\cite{van2008visualizing} method, and have a comparison with the joint-training baseline. For the joint-training baseline, the audio modality is greatly separable, but the vision modality is with poor separability. In contrast, the audio representation separability of our method is ideal, although is slightly worse than the joint-training baseline. And the representation of vision modality has a noticeable improvement. The reason could be that our Diagnosing \& Re-learning strategy can avoid the over-training of well-learnt modality while preserving its discriminative ability, and simultaneously encourage the training of other modalities. These quantitative and qualitative results demonstrate that our method effectively takes into account all modalities during balancing uni-modal learning.

\input{figure_table/figure_cd_acc}

\subsection{Purity and accuracy analysis}
\label{sec:purity_exp}
In our diagnosing process, the learning state of each modality is estimated based on the separability discrepancy between training and validation representation space. And the separability is assessed by the purity of clustering results. In~\autoref{fig:purity-gap}, we record the purity gap between the training and validation representation of the CREMA-D dataset. Based on the results, the audio modality is well-learnt with a huge purity gap while the vision modality is under-fitting. Therefore, during training, the audio encoder tends to be more greatly re-initialized. In experiments, we conduct the proposed method per $20$ epochs for the CREMA-D dataset. The changes in test accuracy during training are shown in~\autoref{fig-acc}. Our method re-initializes the uni-modal encoders, especially one of well-learnt modality. This way firstly results in a sudden drop in performance due to model's reliance on the well-learnt modality, and the performance is progressively recovered and enhanced after re-learning the data. Besides the multimodal accuracy, we also observe the purity of uni-modal test representation. Based on~\autoref{fig:audio-purity} and~\autoref{fig:vision-purity}, for the well-learnt audio modality, the purity of our method also has a similar trend to multimodal accuracy during training. The purity experiences a decreasing and increasing process. And for the under-fitting vision modality, its purity is continuously enhanced during training with our method, which indicates that our method effectively improves its learning.

\subsection{Hyper-parameter sensitivity analysis}
In this section, we conduct experiments to analyze two hyper-parameters $\lambda$ and $H$ in our method. Firstly, when assigning the re-initialization strength based on the uni-modal purity gap, the hyper-parameter $\lambda$ is introduced in~\autoref{equ:strength}, to further control the re-initialization degree. Secondly, our Diagnosing \& Re-learning strategy is conducted per $H$ epoch during training. Here we conduct experiments on both CREMA-D and Kinetics Sounds datasets about these two hyper-parameters. The results are shown in~\autoref{fig:hyper}. For $\lambda$, the results demonstrate that the Kinetics Sounds dataset tends to need a greater re-initialization degree than the CREMA-D dataset, but all these values consistently outperform the joint-training baseline. Also, for the re-initialization frequency $H$, performance is consistently enhanced across different frequencies, and its selection also does not require significant effort.

\input{figure_table/figure_hyper}

%% file: figure_table/table_imbalance.tex
\begin{table*}[t]
\centering
\caption{Comparison with imbalanced multimodal learning methods where bold and underline represent the best and second best respectively. Joint-training is the widely-used baseline with concatenation fusion and one multimodal loss function. }
\label{tab:imbalance}
\setlength{\tabcolsep}{1mm}{
\begin{tabular}{c|cc|cc|cc}
\bottomrule
\multirow{2}{*}{\textbf{Method}} & \multicolumn{2}{c|}{\textbf{\begin{tabular}[c]{@{}c@{}}CREMA-D\\ (Audio/Vision)\end{tabular}}} & \multicolumn{2}{c|}{\textbf{\begin{tabular}[c]{@{}c@{}}Kinetics Sounds\\ (Audio/Vision)\end{tabular}}} & \multicolumn{2}{c}{\textbf{\begin{tabular}[c]{@{}c@{}}UCF-101\\ (RGB/Optical Flow)\end{tabular}}} \\
                                 & \textbf{Acc}                                 & \textbf{Macro F1}                               & \textbf{Acc}                                     & \textbf{Macro F1}                                   & \textbf{Acc}                                   & \textbf{Macro F1}                                \\ \hline
Joint-training                   & 67.47                                        & 67.80                                           & 65.04                                            & 65.12                                               & 80.41                                          & 79.40                                            \\ \hline
G-Blending~\cite{wang2020makes}                       & 69.89                                        & 70.41                                           & {\ul 68.60}                                      & {\ul 68.64}                                         & {\ul 81.73}                                    & {\ul 80.84}                                      \\
OGM-GE~\cite{peng2022balanced}                           & 68.95                                        & 69.39                                           & 67.15                                            & 66.93                                               & 81.15                                          & 80.36                                            \\
Greedy~\cite{wu2022characterizing}                     & 68.37                                        & 68.46                                           & 65.72                                            & 65.80                                               & 80.60                                          & 79.50                                            \\
PMR~\cite{fan2023pmr}                              & 68.55                                        & 68.99                                           & 65.62                                            & 65.36                                               & 81.36                                          & 80.37                                            \\
AGM~\cite{li2023boosting}                              & {\ul 70.16}                                  & {\ul 70.67}                                     & 66.50                                            & 66.49                                               & 81.55                                          & 80.58                                            \\ \hline
Ours                         & \textbf{75.13}                               & \textbf{76.00}                                  & \textbf{69.10}                                   & \textbf{69.39}                                      & \textbf{82.11}                                 & \textbf{80.87}                                   \\ \toprule
\end{tabular}}
\end{table*}

%% file: figure_table/table_gb.tex
\begin{table}[t]
\centering
\caption{Additional comparison with G-Blending~\cite{wang2020makes} method. For a fair comparison, we also introduce the same uni-modal classifier and uni-modal cross-entropy loss function as G-Blending in the Ours\dag \ method.}
\label{tab:gb}
\setlength{\tabcolsep}{1.5mm}{
\begin{tabular}{c|cc|cc|cc}
\bottomrule
\multirow{2}{*}{\textbf{Method}} & \multicolumn{2}{c|}{\textbf{\begin{tabular}[c]{@{}c@{}}CREMA-D\\ (Audio/Vision)\end{tabular}}} & \multicolumn{2}{c|}{\textbf{\begin{tabular}[c]{@{}c@{}}Kinetics Sounds\\ (Audio/Vision)\end{tabular}}} & \multicolumn{2}{c}{\textbf{\begin{tabular}[c]{@{}c@{}}UCF-101\\ (RGB/Optical Flow)\end{tabular}}} \\
                                 & \textbf{Acc}                                & \textbf{Macro F1}                                & \textbf{Acc}                                     & \textbf{Macro F1}                                   & \textbf{Acc}                              & \textbf{Macro F1}                           \\ \hline
Joint-training                   & 67.47                                       & 67.80                                            & 65.04                                            & 65.12                                               & 80.41                                     & 79.40                                       \\
G-Blending~\cite{wang2020makes}                       & 69.89                                       & 70.41                                            & 68.60                                            & 68.64                                               & 81.73                                     & 80.84                                       \\ \hline
Ours                         & 75.13                                       & 76.00                                            & 69.10                                            & 69.39                                               & 82.11                                     & 80.87                                       \\
Ours\dag      & \textbf{79.30}                                       & \textbf{79.58}                                            & \textbf{72.17}                                   & \textbf{72.02}                                      & \textbf{83.05}                            & \textbf{82.17}                              \\ \toprule
\end{tabular}}
\end{table}

%% file: figure_table/table_three.tex
\begin{table}[t]
\centering
\caption{\textbf{(Left):} \textbf{Comparison with imbalanced multimodal learning methods on CMU-MOSI dataset with three modalities.} The greedy method could not extend to cases with more than two modalities. * indicates that the original methods of OGM-GE and PMR only consider two modality cases, but we extend them while retaining the core uni-modal balancing strategy. \textbf{(Right):} \textbf{Comparison with imbalanced multimodal learning methods with Transformer-based backbone on CREMA-D dataset.}} 
\label{tab:three}
\setlength{\tabcolsep}{0.3mm}{
\begin{tabular}{c|ccc|ccc}
\bottomrule
\multirow{2}{*}{\textbf{Method}} & \multicolumn{3}{c|}{\textbf{\begin{tabular}[c]{@{}c@{}}CMU-MOSI\\Transformer-based backbone\\(Audio/Vision/Text)\end{tabular}}} & \multicolumn{3}{c}{\textbf{\begin{tabular}[c]{@{}c@{}}CREMA-D \\Transformer-based backbone\\ (Audio/Vision)\end{tabular}}} \\
                                 & \textbf{Acc}                  & \textbf{Weighted F1}               & \textbf{Macro F1}               & \textbf{Acc}                  & \textbf{Weighted F1}          & \textbf{Macro F1}             \\ \hline
Joint-training                   & 76.96                         & 76.76                              & 75.68                           & 68.55                         & 68.88                         & 69.33                         \\ \hline
G-Blending~\cite{wang2020makes}                       & 77.26                         & 77.20                              & 76.27                           & 69.35                         & 68.77                         & 69.63                         \\
OGM-GE~\cite{peng2022balanced}                           & 77.41*                        & 77.17*                             & 76.09*                          & {\ul 70.70}                   & {\ul 70.67}                   & {\ul 71.18}                   \\
Greedy~\cite{wu2022characterizing}                     & /                             & /                                  & /                               & 68.41 ($\downarrow$)          & 68.72 ($\downarrow$)          & 69.17 ($\downarrow$)          \\
PMR~\cite{fan2023pmr}                              & {\ul 77.55*}                  & {\ul 77.50*}                       & {\ul 76.58*}                    & 69.89                         & 69.70                         & 70.07                         \\
AGM~\cite{li2023boosting}                              & 77.26                         & 77.07                              & 76.02                           & 69.22                         & 69.34                         & 69.61                         \\ \hline
Ours                         & \textbf{77.99}                & \textbf{78.09}                     & \textbf{77.37}                  & \textbf{71.64}                & \textbf{71.66}                & \textbf{72.12}                \\ \toprule
\end{tabular}}
\end{table}

%% file: figure_table/table_noise.tex
\begin{table}[t]
\centering
\caption{Comparison with imbalanced multimodal learning methods on scarcely informative modality case. We modify the audio data of the CREMA-D dataset, adding extra white Gaussian noise to make it noisier and scarcely discriminative. All compared methods have a clear performance drop in this case. In contrast, our method still achieves considerable improvement in this challenging scenario.}
\label{tab:noise}
\setlength{\tabcolsep}{2.5mm}{
\begin{tabular}{c|ccc}
\bottomrule
\multirow{2}{*}{\textbf{Method}} & \multicolumn{3}{c}{\textbf{\begin{tabular}[c]{@{}c@{}}Scarcely informative modality case\\ (Audio/Vision)\end{tabular}}} \\
                                 & \textbf{Acc}                     & \textbf{Weighted F1}             & \textbf{Macro F1}               \\ \hline
Joint-training                   & 65.86                            & 66.03                            & 66.39                           \\ \hline
G-Blending~\cite{wang2020makes}                       & 61.16 ($\downarrow$)              & 60.45 ($\downarrow$)              & 61.15 ($\downarrow$)             \\
OGM-GE~\cite{peng2022balanced}                           & 62.63 ($\downarrow$)              & 64.21 ($\downarrow$)              & 65.07 ($\downarrow$)             \\
Greedy~\cite{wu2022characterizing}                     & 63.17 ($\downarrow$)              & 62.99 ($\downarrow$)              & 63.83 ($\downarrow$)             \\
PMR~\cite{fan2023pmr}                              & {\ul 65.73 ($\downarrow$)}        & {\ul 64.81 ($\downarrow$)}        & {\ul 65.33 ($\downarrow$)}       \\
AGM~\cite{li2023boosting}                              & 62.87 ($\downarrow$)              & 62.28 ($\downarrow$)              & 63.73 ($\downarrow$)             \\ \hline
Ours                             & \textbf{69.49}                   & \textbf{69.72}                   & \textbf{70.28}                  \\ \toprule
\end{tabular}}
\end{table}

%% file: figure_table/figure_tsne.tex
\begin{figure}[t]
\centering
    \begin{subfigure}[t]{0.24\linewidth}
			\centering
			\includegraphics[width=\textwidth]{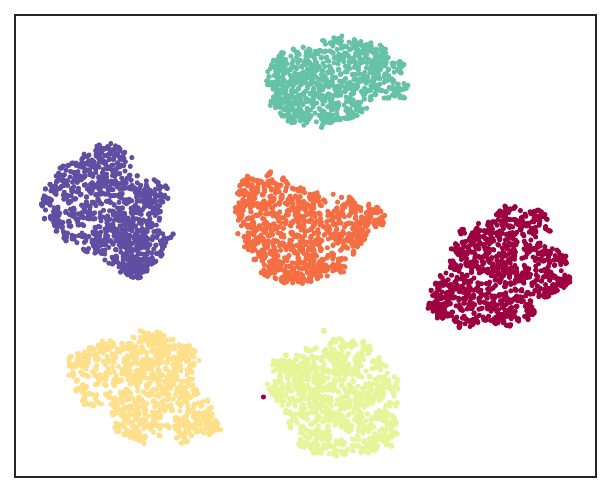}
			\caption{Audio-JT}
			\label{fig-audio-tsne-base}
	\end{subfigure}
	\begin{subfigure}[t]{0.24\linewidth}
			\centering
			\includegraphics[width=\textwidth]{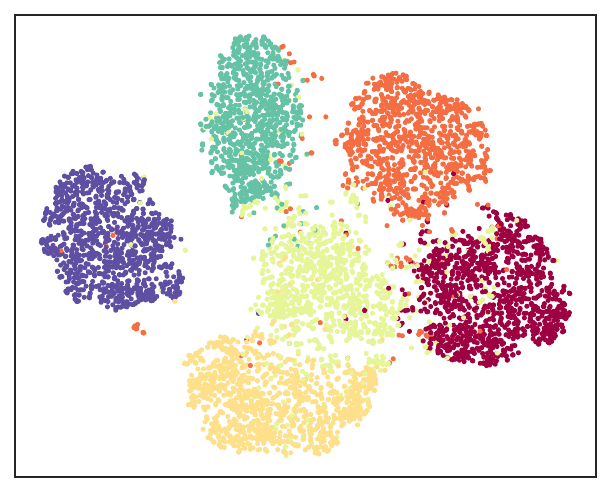}
			\caption{Audio-Ours}
                \label{fig-audio-tsne-our}
	\end{subfigure}
	\begin{subfigure}[t]{0.24\linewidth}
			\centering
			\includegraphics[width=\textwidth]{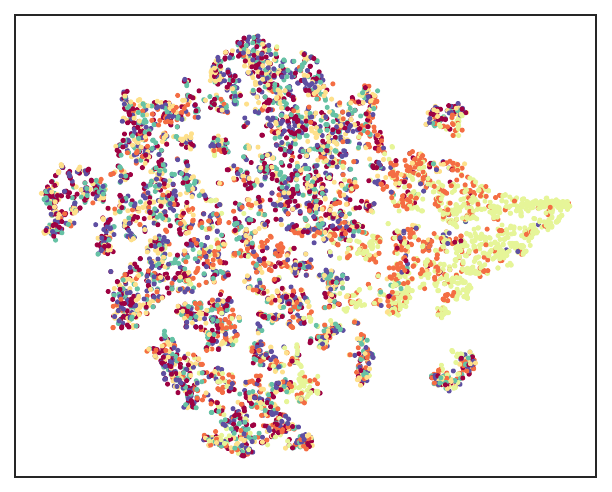}
			\caption{Vision-JT}
                \label{fig-vision-tsne-base}
	\end{subfigure}
	\begin{subfigure}[t]{0.24\linewidth}
			\centering
			\includegraphics[width=\textwidth]{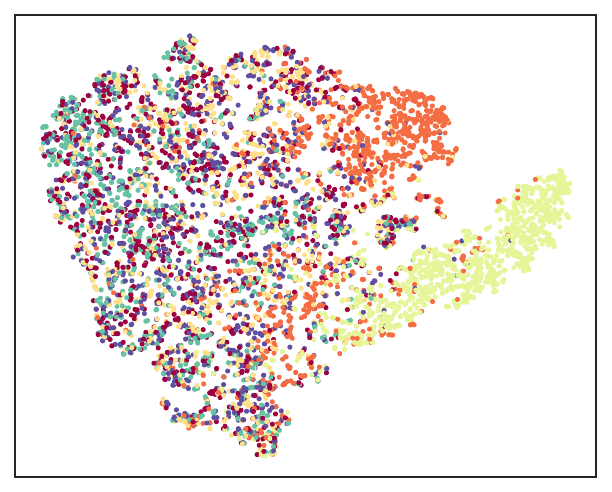}
			\caption{Vision-Ours}
                \label{fig-vison-tsne-our}
	\end{subfigure}
    \caption{Uni-modal representation visualization by t-SNE~\cite{van2008visualizing} on CREMA-D dataset. The categories are indicated in different colors. JT denotes for Joint-training.}
    \label{fig:tsne}
\end{figure}

%% file: figure_table/figure_cd_acc.tex
\begin{figure}[t]
\centering
    \begin{subfigure}[t]{0.24\linewidth}
			\centering
			\includegraphics[width=\textwidth]{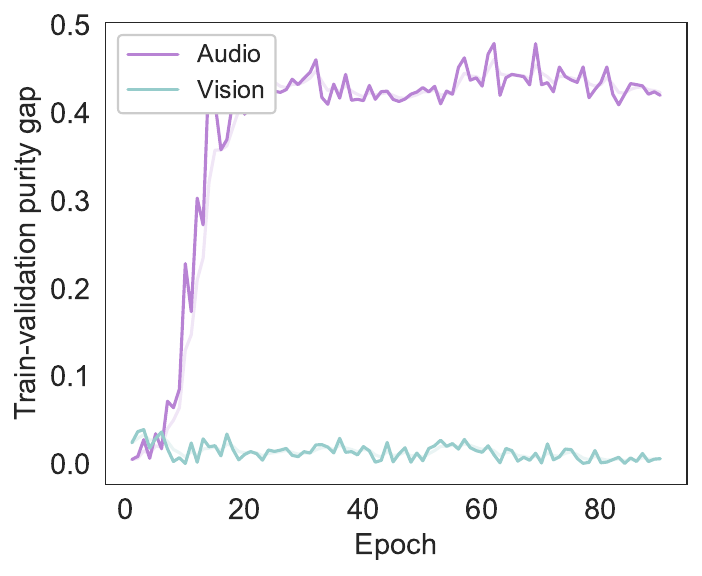}
			\caption{Purity gap.}
			\label{fig:purity-gap}
	\end{subfigure}
    \begin{subfigure}[t]{0.24\linewidth}
			\centering
			\includegraphics[width=\textwidth]{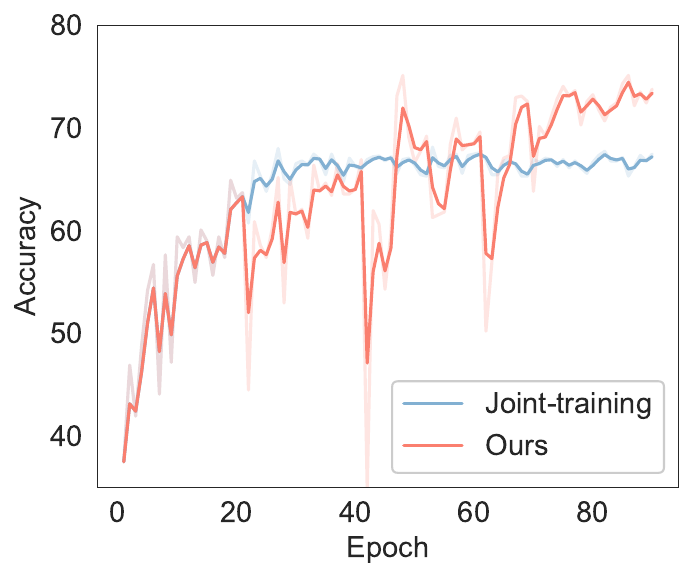}
			\caption{Test accuracy.}
			\label{fig-acc}
	\end{subfigure}
	\begin{subfigure}[t]{0.24\linewidth}
			\centering
			\includegraphics[width=\textwidth]{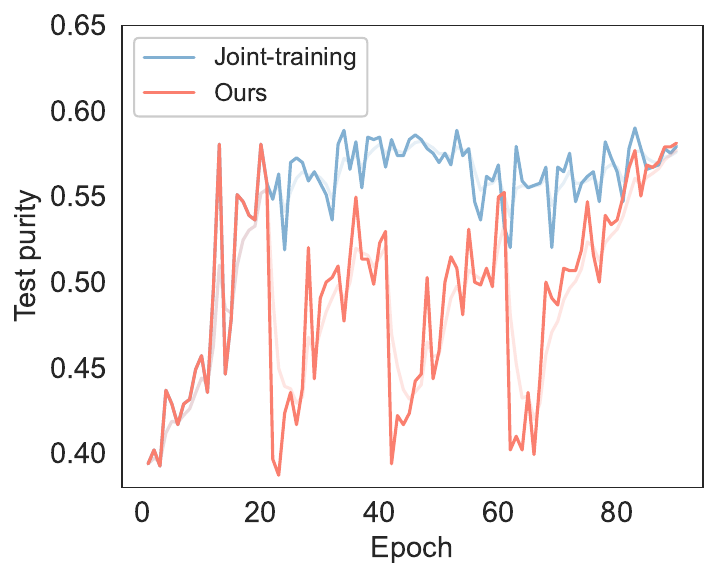}
			\caption{Audio purity.}
                \label{fig:audio-purity}
	\end{subfigure}
	\begin{subfigure}[t]{0.24\linewidth}
			\centering
			\includegraphics[width=\textwidth]{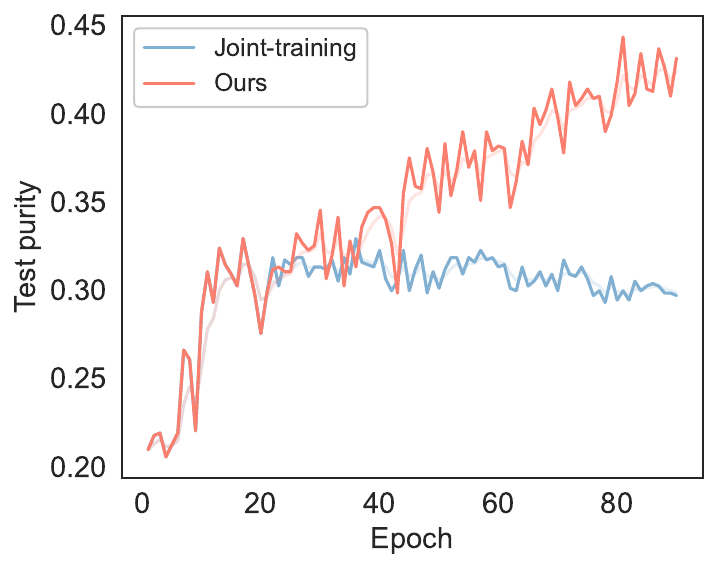}
			\caption{Vision purity.}
                \label{fig:vision-purity}
	\end{subfigure}
    \caption{\textbf{(a):} The purity gap between training and validation representation. \textbf{(b): } Changes in test accuracy during training. \textbf{(c\&d):} The purity of test representation. All results are based on the CREMA-D dataset.}
\end{figure}

%% file: figure_table/figure_hyper.tex
\begin{figure}[t]
\centering
	\begin{subfigure}[t]{0.24\linewidth}
			\centering
			\includegraphics[width=\textwidth]{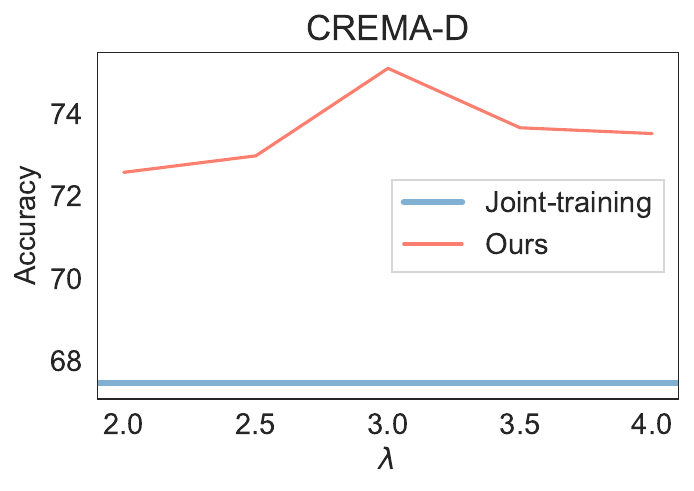}
			\caption{}
	\end{subfigure}
	\begin{subfigure}[t]{0.24\linewidth}
			\centering
			\includegraphics[width=\textwidth]{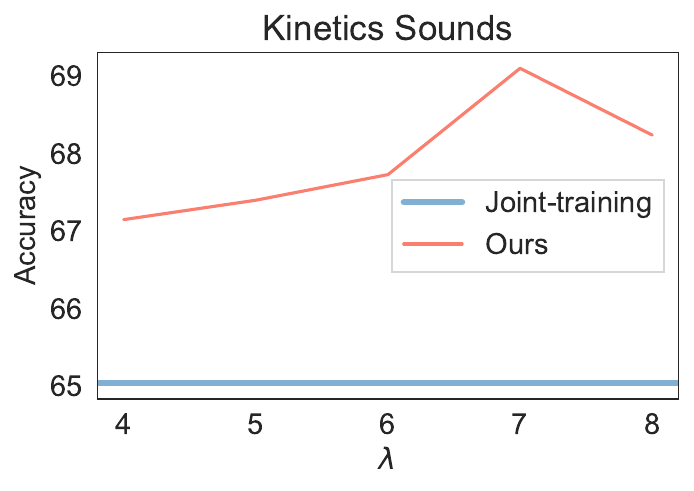}
			\caption{}
	\end{subfigure}
    \begin{subfigure}[t]{0.24\linewidth}
			\centering
			\includegraphics[width=\textwidth]{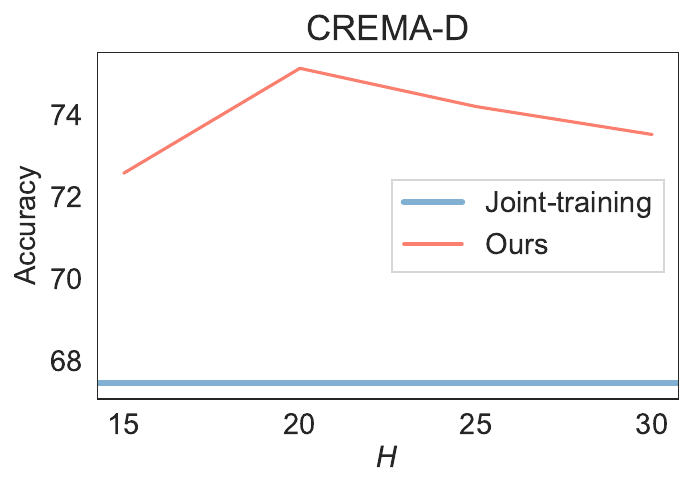}
			\caption{}
	\end{subfigure}
	\begin{subfigure}[t]{0.24\linewidth}
			\centering
			\includegraphics[width=\textwidth]{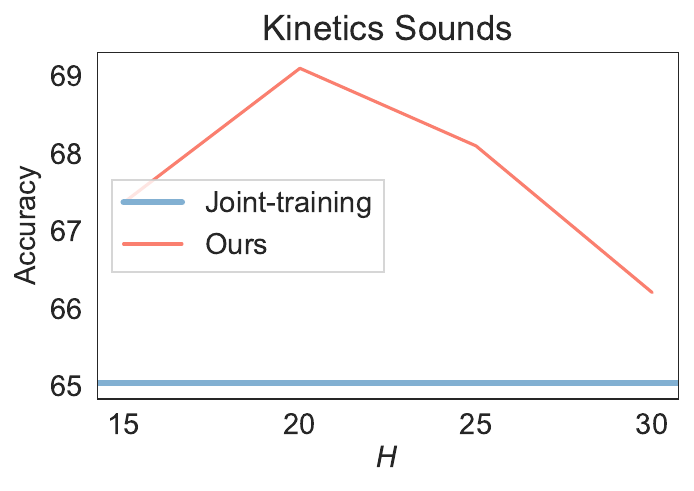}
			\caption{}
	\end{subfigure}
    \caption{Hyper-parameter sensitivity analysis of $\lambda$ in~\autoref{equ:strength} and Diagnosing \& Re-learning frequency $H$ on CREMA-D and Kinetics Sounds datasets.}
    \label{fig:hyper}
\end{figure}

%% file: content/conclusion.tex
\section{Conclusion}
In this paper, we first analyze the limitations of existing imbalanced multimodal learning methods. They ignore the intrinsic limitation of modality capacity and the training of well-learnt modality during balancing. These limitations result in their failure in scarcely informative modality cases and may cause a decrease in the representation quality of well-learnt modality. To this end, we propose the Diagnosing \& Re-learning method. It evaluates uni-modal learning state without any additional modules, and balances uni-modal training by softly re-initializing encoders, benefiting all modalities. Our method not only successfully overcomes the former limitations, but also exhibits its flexibility with diverse multimodal frameworks, well alleviating the imbalanced multimodal learning problem. \\
\textbf{Discussion.} Most current imbalanced multimodal learning methods focus on classification tasks. How to well estimate modality discrepancy and alleviate the imbalance in more types of tasks, \emph{e.g.,} regression tasks, are still under-explored.

%% file: 08072-source.bbl
\begin{thebibliography}{10}
\providecommand{\url}[1]{\texttt{#1}}
\providecommand{\urlprefix}{URL }
\providecommand{\doi}[1]{https://doi.org/#1}

\bibitem{alabdulmohsin2021impact}
Alabdulmohsin, I., Maennel, H., Keysers, D.: The impact of reinitialization on
  generalization in convolutional neural networks. arXiv preprint
  arXiv:2109.00267  (2021)

\bibitem{arandjelovic2017look}
Arandjelovic, R., Zisserman, A.: Look, listen and learn. In: Proceedings of the
  IEEE International Conference on Computer Vision. pp. 609--617 (2017)

\bibitem{ash2020warm}
Ash, J., Adams, R.P.: On warm-starting neural network training. Advances in
  neural information processing systems  \textbf{33},  3884--3894 (2020)

\bibitem{baltruvsaitis2018multimodal}
Baltru{\v{s}}aitis, T., Ahuja, C., Morency, L.P.: Multimodal machine learning:
  A survey and taxonomy. IEEE transactions on pattern analysis and machine
  intelligence  \textbf{41}(2),  423--443 (2018)

\bibitem{cao2014crema}
Cao, H., Cooper, D.G., Keutmann, M.K., Gur, R.C., Nenkova, A., Verma, R.:
  Crema-d: Crowd-sourced emotional multimodal actors dataset. IEEE transactions
  on affective computing  \textbf{5}(4),  377--390 (2014)

\bibitem{fan2023pmr}
Fan, Y., Xu, W., Wang, H., Wang, J., Guo, S.: Pmr: Prototypical modal rebalance
  for multimodal learning. In: Proceedings of the IEEE/CVF Conference on
  Computer Vision and Pattern Recognition. pp. 20029--20038 (2023)

\bibitem{huang2022modality}
Huang, Y., Lin, J., Zhou, C., Yang, H., Huang, L.: Modality competition: What
  makes joint training of multi-modal network fail in deep learning?(provably).
  arXiv preprint arXiv:2203.12221  (2022)

\bibitem{kay2017kinetics}
Kay, W., Carreira, J., Simonyan, K., Zhang, B., Hillier, C., Vijayanarasimhan,
  S., Viola, F., Green, T., Back, T., Natsev, P., et~al.: The kinetics human
  action video dataset. arXiv preprint arXiv:1705.06950  (2017)

\bibitem{li2023boosting}
Li, H., Li, X., Hu, P., Lei, Y., Li, C., Zhou, Y.: Boosting multi-modal model
  performance with adaptive gradient modulation. In: Proceedings of the
  IEEE/CVF International Conference on Computer Vision. pp. 22214--22224 (2023)

\bibitem{liang2021multibench}
Liang, P.P., Lyu, Y., Fan, X., Wu, Z., Cheng, Y., Wu, J., Chen, L., Wu, P.,
  Lee, M.A., Zhu, Y., et~al.: Multibench: Multiscale benchmarks for multimodal
  representation learning. arXiv preprint arXiv:2107.07502  (2021)

\bibitem{liang2022foundations}
Liang, P.P., Zadeh, A., Morency, L.P.: Foundations and recent trends in
  multimodal machine learning: Principles, challenges, and open questions.
  arXiv preprint arXiv:2209.03430  (2022)

\bibitem{van2008visualizing}
Van~der Maaten, L., Hinton, G.: Visualizing data using t-sne. Journal of
  machine learning research  \textbf{9}(11) (2008)

\bibitem{macqueen1967some}
MacQueen, J., et~al.: Some methods for classification and analysis of
  multivariate observations. In: Proceedings of the fifth Berkeley symposium on
  mathematical statistics and probability. vol.~1, pp. 281--297. Oakland, CA,
  USA (1967)

\bibitem{nagrani2021attention}
Nagrani, A., Yang, S., Arnab, A., Jansen, A., Schmid, C., Sun, C.: Attention
  bottlenecks for multimodal fusion. Advances in Neural Information Processing
  Systems  \textbf{34},  14200--14213 (2021)

\bibitem{peng2022balanced}
Peng, X., Wei, Y., Deng, A., Wang, D., Hu, D.: Balanced multimodal learning via
  on-the-fly gradient modulation. In: Proceedings of the IEEE/CVF Conference on
  Computer Vision and Pattern Recognition. pp. 8238--8247 (2022)

\bibitem{qiao2019neural}
Qiao, S., Lin, Z., Zhang, J., Yuille, A.L.: Neural rejuvenation: Improving deep
  network training by enhancing computational resource utilization. In:
  Proceedings of the IEEE/CVF Conference on Computer Vision and Pattern
  Recognition. pp. 61--71 (2019)

\bibitem{sehwag2020separability}
Sehwag, V., Chiang, M., Mittal, P.: On separability of self-supervised
  representations. In: ICML workshop on Uncertainty and Robustness in Deep
  Learning (UDL). vol.~3 (2020)

\bibitem{sokar2023dormant}
Sokar, G., Agarwal, R., Castro, P.S., Evci, U.: The dormant neuron phenomenon
  in deep reinforcement learning. In: Proceedings of the 40th International
  Conference on Machine Learning (2023)

\bibitem{soomro2012ucf101}
Soomro, K., Zamir, A.R., Shah, M.: Ucf101: A dataset of 101 human actions
  classes from videos in the wild. arXiv preprint arXiv:1212.0402  (2012)

\bibitem{wang2020makes}
Wang, W., Tran, D., Feiszli, M.: What makes training multi-modal classification
  networks hard? In: Proceedings of the IEEE/CVF conference on computer vision
  and pattern recognition. pp. 12695--12705 (2020)

\bibitem{wei2024enhancing}
Wei, Y., Feng, R., Wang, Z., Hu, D.: Enhancing multimodal cooperation via
  sample-level modality valuation. In: Proceedings of the IEEE/CVF Conference
  on Computer Vision and Pattern Recognition. pp. 27338--27347 (2024)

\bibitem{wei2024innocent}
Wei, Y., Hu, D.: Mmpareto: boosting multimodal learning with innocent unimodal
  assistance. In: International Conference on Machine Learning (2024)

\bibitem{wei2022learning}
Wei, Y., Hu, D., Tian, Y., Li, X.: Learning in audio-visual context: A review,
  analysis, and new perspective. arXiv preprint arXiv:2208.09579  (2022)

\bibitem{wong2015short}
Wong, K.C.: A short survey on data clustering algorithms. In: 2015 Second
  international conference on soft computing and machine intelligence (ISCMI).
  pp. 64--68. IEEE (2015)

\bibitem{wu2022characterizing}
Wu, N., Jastrzebski, S., Cho, K., Geras, K.J.: Characterizing and overcoming
  the greedy nature of learning in multi-modal deep neural networks. In:
  International Conference on Machine Learning. pp. 24043--24055. PMLR (2022)

\bibitem{xu2023multimodal}
Xu, P., Zhu, X., Clifton, D.A.: Multimodal learning with transformers: A
  survey. IEEE Transactions on Pattern Analysis and Machine Intelligence
  (2023)

\bibitem{yadav2021review}
Yadav, S.K., Tiwari, K., Pandey, H.M., Akbar, S.A.: A review of multimodal
  human activity recognition with special emphasis on classification,
  applications, challenges and future directions. Knowledge-Based Systems
  \textbf{223},  106970 (2021)

\bibitem{yang2024Quantifying}
Yang, Z., Wei, Y., Liang, C., Hu, D.: Quantifying and enhancing multi-modal
  robustness with modality preference. In: The Twelfth International Conference
  on Learning Representations (2024)

\bibitem{ying2019overview}
Ying, X.: An overview of overfitting and its solutions. In: Journal of physics:
  Conference series. vol.~1168, p. 022022. IOP Publishing (2019)

\bibitem{zadeh2016mosi}
Zadeh, A., Zellers, R., Pincus, E., Morency, L.P.: Mosi: multimodal corpus of
  sentiment intensity and subjectivity analysis in online opinion videos. arXiv
  preprint arXiv:1606.06259  (2016)

\bibitem{zaidi2023does}
Zaidi, S., Berariu, T., Kim, H., Bornschein, J., Clopath, C., Teh, Y.W.,
  Pascanu, R.: When does re-initialization work? In: Proceedings on. pp.
  12--26. PMLR (2023)

\bibitem{zhu2021deep}
Zhu, H., Luo, M.D., Wang, R., Zheng, A.H., He, R.: Deep audio-visual learning:
  A survey. International Journal of Automation and Computing  \textbf{18}(3),
  351--376 (2021)

\end{thebibliography}
